\documentclass{article}
\usepackage{eurospeech_01,amssymb,amsmath,epsfig}
\newcommand{\argmax}{\mathop{\rm argmax}}

\newcommand{\p}{{\rm P}}
\ninept
\def\reg{{\rm\ooalign{\hfil
     \raise.07ex\hbox{\scriptsize R}\hfil\crcr\mathhexbox20D}}}

\title{Techniques for effective vocabulary selection}

\makeatletter
\def\name#1{\gdef\@name{#1\\}}
\makeatother
\name{{\em Anand Venkataraman and Wen Wang }}

\address{Speech Technology and Research Laboratory, SRI International \\
Menlo Park, California \\
{\small \tt \{anand,wwang\}@speech.sri.com}
}

\begin{document}
\maketitle
\begin{abstract}

The vocabulary of a continuous speech recognition (CSR) system is a
significant factor in determining its performance.  In this paper, we
present three principled approaches to select the target vocabulary
for a particular domain by trading off between the target
out-of-vocabulary (OOV) rate and vocabulary size.  We evaluate these
approaches against an ad-hoc baseline strategy.  Results are presented
in the form of OOV rate graphs plotted against increasing vocabulary
size for each technique.

\end{abstract}

\section{Introduction}

The size and performance of a language model or speech recognition
system are often strongly influenced by the size of its vocabulary.
Ideally, the vocabulary is small, allowing us to build compact
language models, and it is matched to the target domain so that as
many as possible of the domain-specific words are known to the
recognition system.  Compact language models generate compact word
graphs that are efficient to search and domain-matched vocabularies
result in fewer out-of-vocabulary (OOV) words and consequently
fewer recognition errors.  In a study of the effect of OOV words on
the Word Error Rate (WER) of a recognition system, Rosenfeld
\cite{Rosenfeld:OLN95} arrives at a figure of about 1.2 WER points per
OOV word in a typical large vocabulary task.

While a large and comprehensive vocabulary may be desirable from the
point of view of lexical coverage, we often settle for smaller and
more tractable vocabularies.  Not only are large vocabulary language
models themselves very large, but for speech recognition systems,
there is also the additional cost and effort involved in determining
accurate pronunciations for every vocabulary entry.  Even with the
help of tools to generate pronunciations and check consistency of
entries, this is a difficult task \cite{Lamel96}.

Furthermore, there is also the attendant problem of increased acoustic
confusability for speech recognition systems when the vocabulary is
large \cite{Rosenfeld:OLN95}.  For applications requiring a finite
vocabulary, picking the {\em right}\/ words for the vocabulary is
especially important for achieving satisfactory performance.  Usually,
a number of text corpora from various domains and time periods are
available on which to train.  The target domain is known, and the
amount of data available in the target domain is far less than in any
of the training corpora.  Clearly, restricting the vocabulary to just
the words that are observable in a meager amount of available domain
data would be disastrous.  On the other hand, including the union of
the vocabularies of all the training corpora would be intractable.
What we want in this situation is to assume that the vocabulary of the
target domain is somehow related to the vocabularies of each training
corpus, and subsequently infer the target vocabulary from the
individual training corpus vocabularies, considering the observable
portion of the domain text to be a sample.

Even though vocabulary selection is an important issue and the problem
appears to be simple, little work exists on this topic to date.  The
most common approaches seem to be {\em ad hoc}\/ in nature, typically
including words from each corpus that exceed some threshold frequency.
This threshold depends on intuitions about the relevance of the corpus
to the target domain \cite{Gauvain:Hub498}.  In Rosenfeld's 1995 work 
\cite{Rosenfeld:OLN95} on optimizing vocabularies, attention was
mainly directed at determining the effect on the OOV rate of corpus
recency, size and origin.  While it was found that all three factors
strongly affected the OOV rate, no specific guidelines were proposed
as to how to combine the vocabularies from these different corpora to
choose the target vocabulary.  Indeed, Rosenfeld remarks that an {\em
ad hoc}\/ approach that discounted words by 1\% for every week of age
of the corpus reduced the OOV rate only very slightly for vocabulary
sizes in the range of 20,000 to 50,000 words.

The paucity of work on this important topic can partly be attributed
to the general observation due to Zipf \cite{Zipf:PBL35} that with
even a moderate sized vocabulary chosen wisely, one can hope to get
significant lexical coverage.  Yet it is desirable from the point of
view of scalability, extensibility and generality to study principled
methods to address this problem.  In this paper, we propose three such
principled methods.  The goal is to select a single vocabulary from
many corpora of varying origins, sizes and recencies such that the
vocabulary is optimized for both size and the OOV rate in the target
domain.  Section~\ref{problem} defines the
problem. Section~\ref{method} describes the proposed techniques, and
Section~\ref{results} presents the results.

\section{Problem Description}
\label{problem}

The vocabulary selection problem can be briefly summarized as follows.
We wish to estimate the true vocabulary counts of a partially visible
corpus of in-domain text (which we call the held-out set) when a
number of other fully visible corpora, possibly from different
domains, are available on which to train.  There is an implicit
assumption that the held-out text is related to the training text and
the learning task amounts to inferring this relation.  The reason for
learning the in-domain counts $x_i$ of words $w_i$ is so that the
words may be ranked in order of priority, enabling us to plot a curve
relating a given vocabulary size to its OOV rate on the held-out
corpus.  Therefore, it is actually sufficient to learn some monotonic
function $f(x_i)$ in place of the actual $x_i$.  We may assume that
the counts are normalized by document length so that the amount of
available data for a particular corpus is itself irrelevant to the
issue at hand.

Table~\ref{tbl:problem} illustrates the problem; $n_{i,j}$ are
the visible counts from each of the documents $j$, for the word $w_i$,
and the $x_{i}$ are the incomplete counts for words $w_i$ in the
partially observable domain text.

\begin{table}[htb]
\begin{tabular}{cccccc}\\
{\bf Word} & {\bf Doc 1} & $\cdots$ & {\bf Doc j} & $\cdots$ & {\bf Domain text} \\ \hline
$w_1$      & $n_{1,1}$   & $\cdots$ & $n_{1,j}$   & $\cdots$ & $f(x_{1})$           \\
\multicolumn{6}{c}{$\vdots$}\\
$w_i$      & $n_{i,1}$   & $\cdots$ & $n_{i,j}$   & $\cdots$ & $f(x_{i})$           \\
\multicolumn{6}{c}{$\vdots$}\\
\end{tabular}
\caption{Problem illustration.  We wish to estimate some monotonic
function of the true counts $x_i$ for word $w_i$ in the partially
observed domain text based on a number of fully observed out-of-domain
counts $n_{i,j}$.}
\label{tbl:problem}
\end{table}

Let $x_i$ be some function $\Phi_i$ of the known counts $n_{i,j}$ for
$1 <= j <= m$ for each of the $m$ corpora.  Then, the problem can be
restated as one of learning the $\Phi_i$ from a set of examples where
$$
x_i = \Phi_i(n_{i,1},\cdots,n_{i,m})
$$

In the following section, we summarize three techniques for learning the
$\Phi_i$ that optimize the vocabulary for the domain from which the
held-out data was drawn.

\section{Method}
\label{method}

For simplicity, let the $\Phi_i$ be linear functions of the $n_{i,j}$
and that they are independent of the particular words, $w_i$.  That
is, $\Phi = \Phi_i = \Phi_j, \forall i,j$.  Then, we can write
\begin{equation}
\Phi(n_{i,1},\cdots,n_{i,m}) = \sum_j \lambda_j n_{ij}
\end{equation}

The problem transforms into one of learning the $\lambda_j$.  We now
outline three methods to do this.  The first is based on maximum
likelihood (ML) count estimation, the second and third are based on
document similarity measures.  We evaluate each of these three methods
against a fourth baseline method that simply assigns identical values
to all the $\lambda_j$.

\subsection{Maximum likelihood count estimation}

In ML count estimation, we simply interpret the normalized counts
$n_{ij}$ as probability estimates of $w_i$ given corpus $j$ and the
$\lambda_j$ as mixture coefficients for a linear interpolation.  We
try to choose the $\lambda_j$ that maximize the probability of the
in-domain corpus.  Formally, let $\p(w_i|j) = n_{i,j}$.  Our goal is
to find
\begin{equation}
\hat{\lambda_1},\cdots,\hat{\lambda_m} =
\argmax_{\lambda_1,\cdots,\lambda_m} \prod_{i=1}^{|V|} \left(
\sum_j \lambda_j \p(w_i|j) \right)^{C(w_i)}
\end{equation}
where $C(w_i)$ is the count of $w_i$ in the partially observed
held-out corpus and $V$ is the set of words in the vocabulary.  The
$\lambda_j$ can subsequently be estimated via the EM algorithm
\cite{Dempster:MLF77} and used to calculate the interpolated
normalized counts.  The procedure shown in Figure~\ref{alg:ml-em}, for
instance, is effective in rapidly computing the values of the
$\lambda_j$.
\begin{figure}[htb]
\begin{eqnarray}
\lambda_j &\leftarrow& 1/m \\
\lambda_j' &\leftarrow& \frac
    {\lambda_j \prod_{i=1}^{|V|}\p(w_i|j)^{C(w_i)}}
    {\sum_k \lambda_k \prod_{i=1}^{|V|} \p(w_i|k)^{C(w_i)}} \label{eqn:lambda-reest}\\
\delta &\leftarrow& \lambda_j' - \lambda_j \\
\lambda_j &\leftarrow& \lambda_j' 
\end{eqnarray}
$$
\mbox{Repeat from ~(\ref{eqn:lambda-reest})~ if $\delta >$ some threshold}
$$
\caption{Iterative procedure to calculate the $\lambda_j$.  The
         $\lambda_j$ are reestimated at each iteration until a
         convergence criterion determined by some threshold of
         incremental change is met.  The likelihood of the held-out
         corpus increases monotonically until a local minimum has been
         reached.
}
\label{alg:ml-em}
\end{figure}

\subsection{Document-similarity-based count estimation}

The document-similarity-based count estimation method calculates
interpolation weights from similarity measures between the held-out
corpus and each of the training corpora.  This similarity measure can
presumably be calculated using any of a number of methods ranging from
a simple Euclidean distance metric to a more sophisticated divergence
measure between the observable probability distributions, such as
Kullback-Liebler (KL) \cite{Kullback:IS51} or a symmetric variant
\cite{Jeffreys:IFP46}.

The Euclidean distance metric is calculated as follows.  Suppose we
represent each document by a vector of its normalized word counts.
Then, the Euclidean distance between two corpora $C_a$ and $C_b$,
$\Delta(C_a,C_b)$, is given by
\begin{equation}
\Delta(C_a,C_b) = \sqrt{\sum_{i=1}^{|V|} (n_{a,i} - n_{b,i})^2}
\end{equation}
where $n_{a,i}$ and $n_{b,i}$ are the normalized counts of $w_i$ in
the corpora $a$ and $b$, respectively.

Likewise, the KL-divergence, which we again denote as
$\Delta(C_a,C_b)$ for the sake of uniformity, is given by
\begin{equation}
\Delta(C_a,C_b) = \sum_{i=1}^{|V|} \p(a,i) \log_2 \left(\frac{\p(a,i)}{\p(b,i)}\right)
\label{eqn:kl}
\end{equation}
where we interpret the normalized word counts as probabilities.  

In each of the above distance calculation schemes, let $D_j$ be the
distance of the $j$th corpus from the held-out domain text.  Then,
since the relevance of a corpus to the domain is inversely related to
its distance from the domain, we define
$$
\lambda_j = \frac{1/D_j}{\sum_k 1/D_k}
$$

\subsection{Data sources and implementation}

The experimental setup consisted of learning the optimal vocabulary to
model the language of the English broadcast news.  A small amount of
hand-corrected closed captioned data, amounting to just under 3
hours (about 25,000 words), drawn from six half-hour broadcast
news segments from January 2001, was used as the {\em partially
visible}\/ held-out data to estimate the two mixture weights
$\lambda_1$ and $\lambda_2$.  This held-out data is part of the corpus
released by the Linguistic Data Consortium (LDC) for the National
Institute of Standards and Technology (NIST) sponsored English topic
detection and tracking (TDT4) task.

The training corpora were deliberately chosen to be as different from
each other in character as possible.  The first corpus consisted of
about 18.5 million words of English newswire data covering the period
July 1994 through July 1995, and was distributed by the LDC for the
NIST-sponsored Hub3 1995 continuous speech recognition task.  It
contained text from The NY Times News Service, LA Times,
Washington Post News Service, Wall Street Journal and Reuters North
American Business News.  The second training corpus consisted of a
closer match to the target domain and came from segments of the TDT4
dataset released by the LDC.  This consisted of about 2.5 million
words of closed captioned transcripts from the period November through
December 2000.

Unigram counts for the training and held-out corpora were generated
using language modeling tools from the SRILM \cite{Stolcke:SRILM}
using Witten-Bell \cite{Witten:ZFP91} smoothing.  Estimation of the
$\lambda_j$ was performed on five of the six held-out segments which
we collectively refer to as the development corpus, and OOV rates were
measured on the remaining segment, which we refer to as the test
corpus.  This procedure was repeated six times, one for each possible
split of the held-out data.  The results we present are averaged
numbers obtained from the six splits.  Where applicable, we use the
subscripts ``hub3'' and ``tdt4'' to refer to parameters specific to
the above corpora.

\section{Results and Discussion}
\label{results}

\begin{figure}[htb]
\begin{center}
\includegraphics[width=3.25in,height=3in]{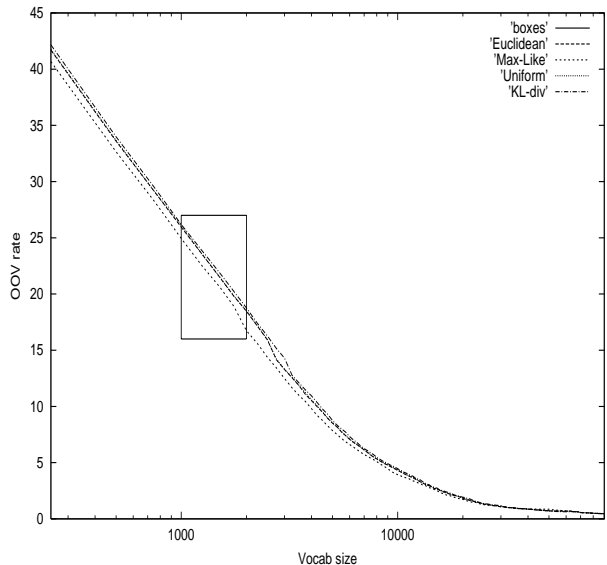}
\vspace{10pt}
\caption{Averaged OOV rate across the six test corpora
  when the vocabulary was determined by each of the four methods
  described in this paper.  This plot is meant for the purpose of
  depicting the general trend.  Expansions of the rectangular
  enclosure in a subsequent plot will serve as a more detailed point of
  discussion.
  }
\label{fig:oov-1k-90k}
\end{center}
\end{figure}

We examine the results of our experiments to evaluate the various
methods.  Figure~\ref{fig:oov-1k-90k} shows a plot of the OOV rate
against increasing vocabulary size from 1 word to 90,000 words.  This
figure, which is plotted in the logarithmic scale, is only meant to
show the general shape of the individual plots and for drawing some
broad generalizations.  For instance, we see confirmation of the
common observation that the OOV rate of a given vocabulary on a corpus
is logarithmically related to the vocabulary size.  Furthermore, it is
also evident that for small vocabularies there exist obvious
differences in the performances of the various vocabulary selection
methods.  But for large vocabularies, this difference is seen to
diminish.  Indeed, for vocabulary sizes in excess of about 60,000
words, the four plots practically merge into a single line showing
that at around that threshold and beyond, we capture practically all
the words that are likely to be used in the domain under
consideration, regardless of the specific method used to choose the
vocabulary.

\begin{figure}[htb]
\begin{center}
\includegraphics[width=3.25in,height=3in]{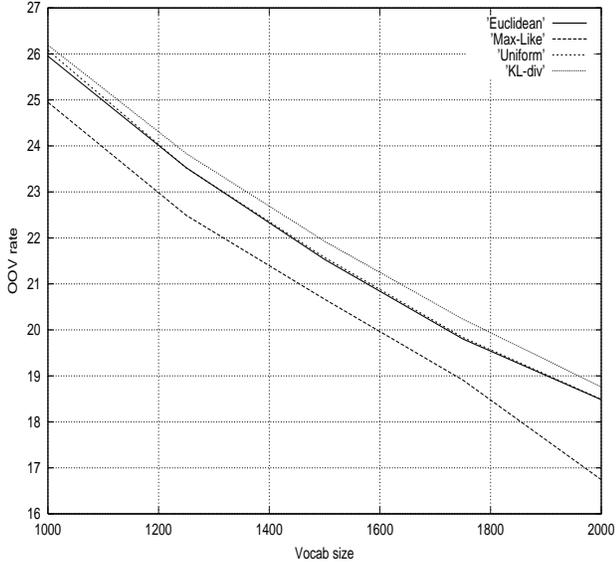} 
\vspace{10pt}
\caption{ Averaged OOV rate across the six test corpora
  when the vocabulary was determined by each of the four methods
  described in this paper.  The plot shows the segment of the OOV rate
  curve for a vocabulary size in the range of 1,000 to 2,000 words.
  }
\label{fig:oov-1k-2k}
\end{center}
\end{figure}

For a finer-grained comparison of the individual techniques, we
restrict our attention to the rectangular sub-region in
Figure~\ref{fig:oov-1k-90k}, which is depicted in a separate plot in
Figure~\ref{fig:oov-1k-2k}.  It shows the performance of the four
systems for a vocabulary range of 1,000 to 2,000 words.  The trend of
the curves in this graph, which continues up to a vocabulary size of
around 40,000 words, clearly shows that the ML method outperforms all
the other three methods by over 1\% absolute.  It is also clear that
the method based on KL-divergence is the poorest of all, performing
worse than even the uniform baseline.  The Euclidean-distance-based
method performs almost identically as the uniform baseline (and thus
the plot for the latter, being almost hidden behind that of the
former, is barely visible).

In hindsight, the relatively good performance of the
maximum-likelihood-based method is not very surprising because it is
the only method that does not {\em look beyond}\/ the development
corpus vocabulary to compute its objective function.  Both the
KL-divergence-based method and the Euclidean-distance-based method sum
quantities over the entire vocabulary and are therefore affected by
the values held by individual words that were not seen in the
development corpus.  This problem is especially acute because the
actual vocabulary of the partially visible development corpus is
typically tiny compared to the vocabularies of the training corpora.
The KL-divergence-based method is affected most by this situation.
Because KL-divergence involves calculation of log-probabilities, the
method is extremely sensitive to the amount of probability mass
devoted to unseen vocabulary items and consequently to the particular
form of smoothing employed.  Since a significant number of words in
the vocabulary are typically unseen in the development corpus, these
end up with very low unigram probabilities.  Thus, in summing over the
entire vocabulary, large negative numbers come into play which
overshadow any significant contribution to the total divergence by the
unigrams observed in the development corpus.  We suspect therefore
that we must not attach much significance to the final quantity
computed by this method unless the size of the development corpus
itself is substantial.

The Euclidean method is also likewise affected, but to a lesser degree
and slightly differently.  The computed distances tend to be dominated
by words that are absent in the development corpus rather than by words
that are present in it.  Since the absent words form the bulk of the
vocabulary, the distances computed between the various corpora and the
development text, and consequently the $\lambda_j$ will all roughly be
the same, as evidenced by the figures in Table~\ref{tbl:lambdas}.
\begin{table}[htb]
\begin{center}
\begin{tabular}{lllll}
{\bf Method} & $\lambda_{\mbox{tdt4}}$ & $\lambda_{\mbox{hub3}}$ & $\Delta_{\mbox{tdt4}}$ & $\Delta_{\mbox{hub3}}$ \\ \hline 
Max Like      &  0.89 & 0.11   & n/a   & n/a   \\
Euclidean     &  0.51 & 0.49   & 1.36  & 1.44  \\
KL-Div        &  0.42 & 0.58   & 92.86 & 66.09 \\
Uniform       &  0.50 & 0.50   & n/a   & n/a   \\
\end{tabular}
\vspace{10pt}
\caption{Inferred interpolation weights $\lambda_j$, along
with the normalized corpus distances from the domain text for the
distance-based methods.  All figures are averaged across all six
splits of the test data.
}
\label{tbl:lambdas}
\end{center}
\end{table}

\section{Conclusions}
\label{conclusion}

We have outlined three general techniques to select an optimal
vocabulary for domain-specific speech and language modeling tasks.
The techniques are scalable to arbitrarily large-sized corpora and
extensible to any number of corpora.  Whenever reasonable amounts of
training data and reliable unigram count estimates are available, we
believe that the maximum-likelihood-based method we have described is
a robust way to select a domain's vocabulary especially when its size
is expected to be under a certain threshold.  This threshold can be
expected to vary between domains and it is possible that when it is
high, the choice of any particular strategy over another does not
matter.  However, we believe that always following a principled
strategy to select the vocabulary offers the safest path.

We plan to continue to refine and evaluate the techniques presented in
this paper and apply them for vocabulary selection in the English
broadcast news recognition task of the NIST 2003 Hub4 evaluation.

\section{Acknowledgments}

This material is based upon work supported by the Defense Advanced
Research Projects Agency under Contract No. MDA972-02-C-0038.  The
authors also thank Andreas Stolcke and Kristin Precoda for useful
discussions regarding its content and Judith Lee for careful
proof-reading of the manuscript.

\bibliographystyle{ieee}
\bibliography{vocab-selection}

\end{document}